\pdfoutput=1

\documentclass[11pt]{article}

\usepackage{acl}

\usepackage{booktabs}

\usepackage{times}
\usepackage{latexsym}
\usepackage{multirow}
\usepackage{amsmath}
\usepackage{listings}
\usepackage{graphicx}

\usepackage[T1]{fontenc}

\usepackage[utf8]{inputenc}

\usepackage{microtype}

\usepackage{color} 

\title{Peering into the Mind of Language Models: An Approach for Attribution in Contextual Question Answering}

\author{Anirudh Phukan, Shwetha Somasundaram, Apoorv Saxena, Koustava Goswami \\ {\bf Balaji Vasan Srinivasan} \\
        Adobe Research, India
\\  \{phukan, shsomasu, apoorvs, koustavag, balsrini\}@adobe.com}

\begin{document}
\maketitle
\begin{abstract}
With the enhancement in the field of generative artificial intelligence (AI), contextual question answering has become extremely relevant. Attributing model generations to the input source document is essential to ensure trustworthiness and reliability. We observe that when large language models (LLMs) are used for contextual question answering, the output answer often consists of text copied verbatim from the input prompt which is linked together with "glue text" generated by the LLM. Motivated by this, we propose that LLMs have an inherent awareness from where the text was copied, likely captured in the hidden states of the LLM. We introduce a novel method for attribution in contextual question answering, leveraging the hidden state representations of LLMs.  Our approach bypasses the need for extensive model retraining and retrieval model overhead, offering granular attributions and preserving the quality of generated answers. Our experimental results demonstrate that our method performs on par or better than  GPT-4 at identifying verbatim copied segments in LLM generations and in attributing these segments to their source. Importantly, our method shows robust performance across various LLM architectures, highlighting its broad applicability. Additionally, we present \textsc{Verifiability-granular}\footnote{Dataset is available at \url{https://github.com/Anirudh-Phukan/verifiability-granular}.}, an attribution dataset which has token level annotations for LLM generations in the contextual question answering setup.

\end{abstract}

\begin{figure*}[h!]
    \centering
    \includegraphics[width=\linewidth]{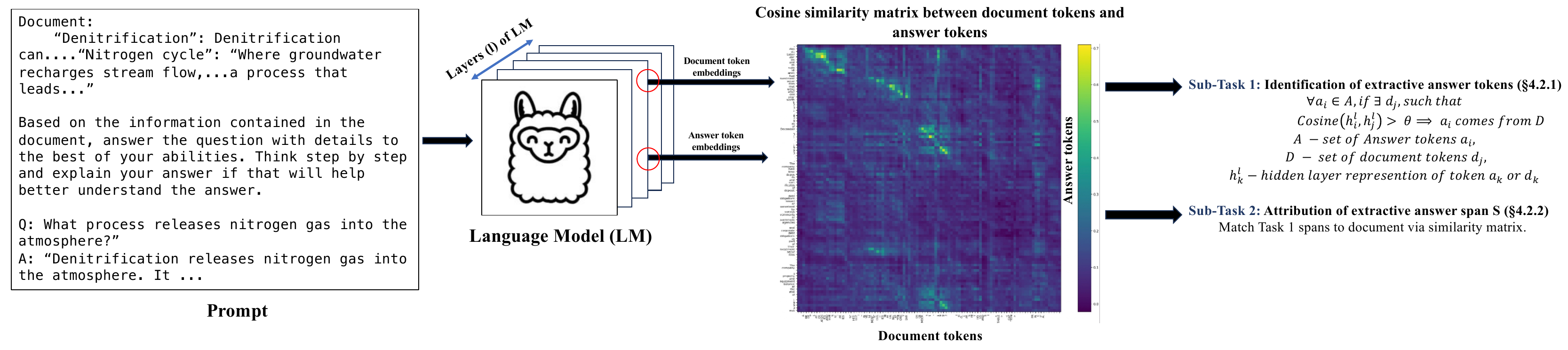}
    \caption{Overall picture of our proposed methodology; our method utilizes
    hidden layer representations from both the document and answer to determine the attribution of answer tokens. Initially, we identify
    extractive answer tokens (\S\ 
 4.2.1) through a cosine similarity matrix between document and answer tokens. Subsequently, we map these tokens to document token sequences by identifying anchor tokens and generating  candidates, later ranked based on their cosine similarity to achieve attribution. (\S\ 
 4.2.2)}
   
    \label{fig:pipeline}
\end{figure*}

\section{Introduction}
The surge in the capabilities of Large Language Models (LLMs) has revolutionized natural language understanding. Their ability to comprehend and generate human-like text has resulted in their widespread adoption across various industries. A prominent and pivotal application of these models is question answering, particularly in contextual settings. The ability of LLMs to parse, interpret, and respond to queries within a given context has facilitated efficient information retrieval and comprehension.

Despite the remarkable strides made in contextual question answering, challenges persist within LLMs. While LLMs excel in generating informative responses, they often fall short in providing explicit references or attributions to the specific sections or sources within the context from which their answers derive \cite{liu2023evaluating}. The absence of this attribution impedes the ability to verify the authenticity and accuracy of the generated information. Attribution not only enables verification \cite{rashkin2023measuring} but also increases user trust and confidence in the responses generated by these models, thus fostering their broader acceptance and utilization across diverse domains \cite{bohnet2022attributed}.

Existing methods for attribution typically align within three primary categories \cite{Li2023ASO}: Systems fine-tuned or trained explicitly for attribution tasks \cite{weller2023according, gao2023enabling}, retrieve-then-read \cite{chen2017reading, lee2019latent} and approaches relying on post-generation attribution \cite{gao2023rarr, huo2023retrieving}. However, each of these approaches encounters substantial challenges in achieving accurate and granular attribution, thereby limiting their effectiveness.

The initial category, which consists of systems tailored or explicitly trained for attribution tasks, poses a significant challenge \cite{huang2023citation}. This challenge primarily arises because the training process demands extensive resources, including time, computational power, and a large corpus of annotated data.

Furthermore, the continuous development of more sophisticated models requires retraining, perpetuating an unending cycle of adaptation. Additionally, the need for regression testing to ensure answer quality amplifies the burden, creating a considerable bottleneck in practical application and scalability.

On the other hand, the second category—post-generation attribution—offers an alternative but not without its own set of drawbacks. This approach relies on leveraging retrieval models to trace the sources of the generated answers. However, this method suffers from the overhead associated with these retrieval models, which often demand substantial computational resources and time. Furthermore, the granularity of attribution is constrained by the chunk size at which retrieval is performed, limiting the precision of attribution.

Lastly, the third category—retrieve then read methods—also grapples with its unique set of complications. Similar to the post-generation attribution approach, this method faces the problem of retrieval method overhead and chunking issues. The concept here is to utilize retrieved evidence as a basis for generating answers and consequently, attributions. However, as outlined in \cite{gao2023rarr}, retrieval does not equate to attribution due to the potential integration of external knowledge. 

Our proposed method aims to surmount the limitations posed by existing attribution approaches by adopting a distinct strategy that eliminates the need for additional training while offering granular attributions. By delving into the inner workings of the LLM during its generation process, we sidestep the resource-intensive training requirements.

The crux of our approach lies in accessing the hidden state representations of tokens produced by the LLM when generating responses. Leveraging the contextual cues provided by the input—context, question, and generated answer—we extract these hidden state representations via a forward pass through the model. These hidden representations which happen to be contextual embeddings of tokens are then matched to perform attribution. An overview of this process is shown in Figure \ref{fig:pipeline}.

Operating at a token-level granularity mirrors the natural generation process of LLMs. Unlike conventional post generation attribution methods that grapple with decisions on chunking at paragraph or sentence levels, our approach bypasses this dilemma. By steering clear of chunk-based attributions, we are liberated from arbitrary segmentation of the context, which often leads to imprecise referencing and diluted contextual connections.

The existing public datasets curated for the task of attribution (attribution of LLM generations) have annotations only at the response level or at the sentence level (for every sentence within the response). \cite{kamalloo2023hagrid, liu2023evaluating, Malaviya2023ExpertQAEQ}. To assess our method's effectiveness, token-level annotations are essential. Consequently, we process the data collected by \citet{liu2023evaluating} to generate token-level annotations.
We release this dataset to facilitate further research on token-level attribution.

In essence, our approach offers a unique and efficient solution to the problem of attribution in contextual question answering. Additionally, we also introduce a dataset that facilitates token-level attribution. With this in mind, we are excited to share our primary contributions in this paper.

\begin{itemize}
    
    \item We introduce a pioneering method for attribution in contextual question answering. The strengths of our approach include its token-level granularity ensuring precise attributions, lack of additional retrieval model overhead and its training-free nature which ensures consistent answer quality.
    
    \item Our experimental findings demonstrate the efficacy of our method across various model families. This indicates that leveraging hidden layer representations for attribution can be broadly applied across different LLM architectures, highlighting the wide-ranging applicability of our approach.
    
    \item Additionally, we release the \textsc{Verifiability-granular} dataset, which contains token-level attributions for LLM generations in the contextual question answering setup.
\end{itemize}

\section{Related Work}

Researchers have employed different approaches for the attribution of generated or identified text spans. A shared task was organized by \citet{DBLP:journals/corr/abs-1811-10971}  to encourage researchers to build systems capable of performing fact verification and attribution to the source texts. \citet{DBLP:journals/corr/abs-2110-06674} emphasized the role of source attribution in fostering truthful and responsible AI. The growing concern of fake news detection in AI-based news generation has also been considered as an attribution task \cite{pomerleau2017fake,DBLP:conf/naacl/FerreiraV16}. In addition to automatic attribution, studies on manual attribution have been performed by domain specific individuals \cite{borel2023chicago,DBLP:journals/sigkdd/LiGMLSZFH15}. While we discuss about fact verification as an attribution task, it is important to note that user interactions have also been found to require attribution \cite{dziri-etal-2022-evaluating}. \citet{DBLP:journals/corr/abs-2207-06220} has highlighted the fact that wikipedia articles needs validation and that the citations need to be attributed. On the other hand \citet{sarti-etal-2023-inseq} introduced a python library based on GPT-2 capable of identifying feature attributions generated from the Insequence model; capable of performing in multilingual settings \cite{DBLP:journals/corr/abs-2310-01188}. Recently, researchers performed attribution task on multimodal systems as well \cite{DBLP:journals/corr/abs-1711-06104,DBLP:journals/inffus/HolzingerMSP21,DBLP:conf/emnlp/ZhaoCWJLQDGLLJ23}.

However, what is missing in prior art is the utilisation of the contextual guidance inherently present during generation for the attribution task. In our work, we tackle this problem by utilising the contextual information encoded during generation by the LLM and attribute spans to semantically relevant parts of the source document.

\begin{figure}
    \centering
    \includegraphics[width=1\linewidth]{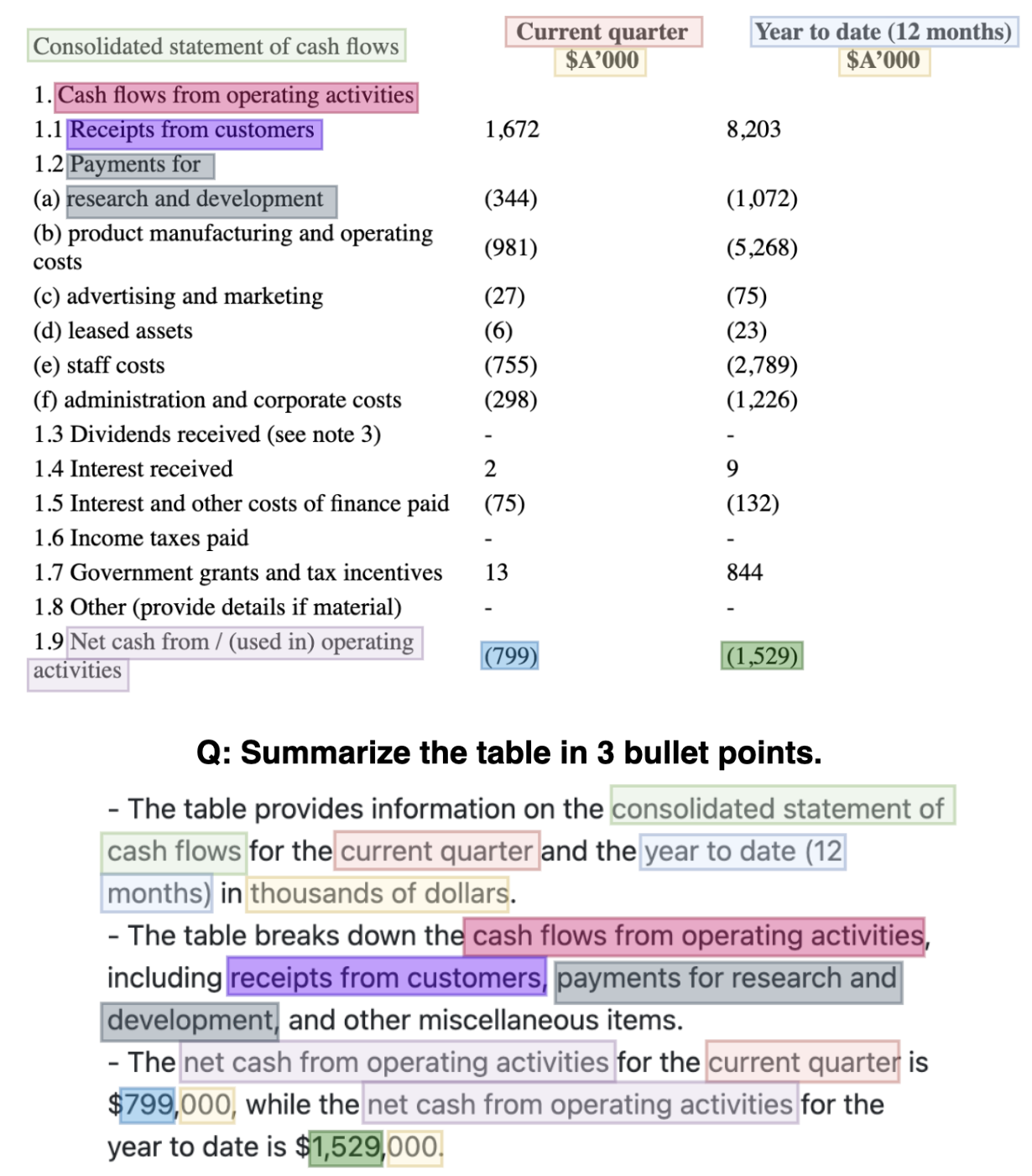}
    \caption{Semi Extractive answers by LLMs}
    \label{fig:SemEx}
\end{figure}

\section{Problem Statement}

We observe a phenomenon in contextual question answering using LLMs which results in a comprehensive answer typically characterized by factual spans, replicated verbatim from various segments of the provided context, interwoven with "glue text". An example of this phenomenon is shown in Figure \ref{fig:SemEx}. \citet{yang2023inference} also note this pattern and use references to losslessly speed up LLM inference. 

Building upon the aforementioned observations, we dissect the challenge of attribution in the contextual generation setting into two distinct yet interconnected sub-problems. The first sub-problem involves the identification of tokens within the output that have been directly copied from the provided context. 

The second sub-problem delves deeper into the attribution of these identified tokens. This step involves mapping these tokens back to their original positions within the document. In essence, this implies tracing back the tokens to the exact sections within the document from which they were copied verbatim. 
 
By doing this, we aim to establish a clear flow of information from the source document to the generated output, thereby enabling a more nuanced understanding of the context generation process.

The task at hand involves two distinct subtasks within the context of attributing the output of a large language model (LLM). We are given a document $D$ represented as a sequence of tokens $(d_1, d_2, \ldots, d_n)$, a question $Q$, and an answer $A$ represented as a sequence of tokens $(a_1, a_2, \ldots, a_m)$. Our objective comprises two interconnected subtasks:

\textbf{Subtask 1: Token Attribution Identification.} This subtask involves identifying a subset of tokens in $A$, denoted as $A_{\text{attr}} \subseteq A$, that require attribution to tokens in $D$. Mathematically, this can be expressed as selecting tokens $a_i \in A$ that are directly influenced or copied from $D$. In our current study, we restrict the scope to tokens that are verbatim copied from $D$.

\textbf{Subtask 2: Corresponding Token Mapping.} For each token in $A_{\text{attr}}$, this subtask aims to find a mapping function $f: A_{\text{attr}} \rightarrow D$ such that for every token $a_i \in A_{\text{attr}}$, there exists a corresponding token or sequence of tokens in $D$ to which $a_i$ can be attributed. 

Formally, the mapping function can be expressed as:

\begin{align*}
&f: A_{\text{attr}} \rightarrow D, \quad \text{s.t. } \forall a_i \in A_{\text{attr}}\\
&f(a_i) = (d_j, \ldots, d_k) \iff \\
&a_i \text{ is attributed to } (d_j, \ldots, d_k) \text{ in } D
\end{align*}

In this formulation, the function $f$ establishes a connection between the tokens in the answer requiring attribution and their corresponding tokens or token sequences in the document $D$.

\section{Proposed Work}

\subsection{Motivation}
Humans easily identify which parts of a document they use to answer questions. This skill to trace information back to its source is key for effective communication, especially when discussing complex ideas. Similarly, we propose that Large Language Models (LLMs) have an inherent awareness of the document parts they use while generating answers. This awareness is likely captured within the hidden states of the LLM, which encode token information during the generation process.

Our work is based on the idea that if LLMs can generate responses combining copied segments and self-generated "glue text", they must inherently differentiate between copied content and self-generated content. This differentiation is what we aim to uncover. By accessing and analyzing the hidden states of an LLM, we seek to reveal how it uses and attributes source information in its responses. 

This approach allows us to trace how the LLM processes and uses the provided context at a detailed, token-level granularity. We believe that fully understanding how LLMs generate responses requires examining the process at this fundamental level - where each token's role in the response is clarified and attributed to its original source. Not only does this give us an understanding of a LLM's inner workings, but also accomplishes the useful task of attributing outputs, which helps humans easily verify/navigate the LLM outputs.

\subsection{Methodology}
\label{method}

Given a language model, denoted as $M$, we construct a prompt $P$ by concatenating the document $D$, represented as a sequence of tokens $(d_1, d_2, \ldots, d_n)$, the question $Q$, and the answer $A$, represented as a sequence of tokens $(a_1, a_2, \ldots, a_m)$. This concatenation is formally expressed as $P = D + Q + A$, where `+` denotes the concatenation operation. The prompt $P$ is then passed through the model $M$ in a forward pass to obtain the hidden layer representations for each token in $P$. These representations capture the contextual information encoded by the model for each token. An illustrative example of this process is provided below.

\begin{lstlisting}[basicstyle=\small\ttfamily, breaklines=true, breakatwhitespace=true]
[INST]
Document:
{document}
Based on the information contained in the document, answer the question with details to the best of your abilities. Think step by step and explain your answer if that will help better understand the answer. 
Q: {question} A:
[/INST]
{answer}
\end{lstlisting}

Let's denote the hidden layer representation of each token $t_i$ for a specific layer $l$ as $h_i^l$.

\subsubsection{Identifying extractive output tokens}

For the first sub-task of identifying tokens in answer $A$ that are directly copied from document $D$, we perform the following operation. For any specific layer $l$ and for all tokens $a_i \in A$, we say that a token $a_i$ comes from $D$ if there exists a token $d_j \in D$ such that the cosine similarity between $h_i^l$ and $h_j^l$ is greater than a threshold $\theta$.

This can be formally represented as:
\begin{align*}
& \forall a_i \in A, \quad \text{if } \exists d_j \in D \text{ s.t. } \\
& \text{Cosine}(h_i^l, h_j^l) > \theta \implies a_i \text{ comes from } D
\end{align*}

\subsubsection{Attributing Extractive Spans}

Given a span $S$ in $A$ to be attributed, consisting of tokens $a_1, \dots, a_n$, we compute the average hidden layer representation $h_S$ for each token $a_i \in S$ as:

\begin{align*}
h_S = \frac{1}{n} \sum_{i=1}^{n} h_i^l
\end{align*}

Next, we use $h_S$ to identify anchor tokens in $D$. For each token $d_j \in D$, we compute the cosine similarity between $h_S$ and $h_j^l$ and select tokens with the highest similarities as anchor tokens, denoted as $D_T$. 

For each anchor token $d_a \in D_T$, we explore windows of tokens around $d_a$, up to a maximum length $L$. We calculate the average hidden layer representation $h_W$ for each window $W$ and identify the window with the highest similarity to $h_S$. The highest-ranked window is considered the final attribution for the span $S$.

In cases where $D$ is segmented into evidence spans $e \in E$, the score for each $e$ is the similarity between $h_S$ and the best window within $e$.

\section{Experimental Setup}
In this section, we outline the details of our experimental setup; the dataset, the evaluation metrics; and the baselines.
\subsection{Datasets}
\label{section:data}

\textbf{QuoteSum} \cite{schuster2023semqa}\textbf{:}\footnote{Downloaded from \url{https://github.com/google-research-datasets/QuoteSum/tree/main}. The dataset is licensed under CC-BY-SA-4.0 license.} This dataset consists of questions, relevant passages, and human-written semi-extractive answers. In the process of dataset construction, the human annotators were tasked to answer multiple source questions by combining information from various sources. They were instructed to explicitly extract factual spans and weave them together into a coherent, well-grounded passage. The dataset comprises of 4,009 semi-extractive answers to 1,376 unique questions in total and is split into train, validation and test sets with ratios 60\%, 7\%, 33\%, respectively.

\textbf{Verifiability-Granular:} We also curate a dataset with token-level attributions called \textsc{Verifiability-granular (Veri-gran)} by processing the dataset introduced by \cite{liu2023evaluating}\footnote{Downloaded from \url{https://tinyurl.com/verifiability}. The dataset is licensed under MIT license.}. The original dataset contains commercial generative search engines generated responses to input queries along with the retrieved content used for the generation. Annotations map sentences in the response to portions of the source text.

We split the source text using \textit{nltk.sent\_tokenize}\footnote{\url{https://github.com/nltk/nltk}} and filter the samples that map a response statement to only one sentence of the source text. Characters are matched using \textit{diff\_match\_patch}\footnote{\url{https://github.com/google/diff-match-patch}} and tokens are annotated based on if all characters in the corresponding tokens match. Some additional post-processing (removing spans containing only punctuation and stop words) is done to obtain annotations in the same format as the QuoteSum dataset. The dataset contains 170, 197 annotated statements and 272, 320 annotated spans in the dev and test set respectively.

\subsection{Metrics}

The evaluation metrics for the two sub-tasks are described separately below.

\subsubsection{Metrics for Sub-task 1}

For the first sub-task, which involves the identification of spans that have been directly copied from the context, we use token-level precision, recall, and F1 score as our metrics. The ground truth for this task is the set of tokens marked as spans explicitly extracted from the context.

\subsubsection{Metrics for Sub-task 2}

For the second sub-task, which entails attributing the identified spans to their original positions within the document, we use accuracy as our metric. Accuracy is computed as the fraction of instances where the system correctly predicts the paragraph from which the span was extracted. This allows us to measure how effectively the system can trace the origin of the spans within the document.

\subsection{Baselines}

In the context of our study, we identify and utilize a number of baselines for both sub-tasks.

\subsubsection{Baselines for Sub-task 1}

For the first sub-task, the identification of copied spans, it is worth noting that, to our best knowledge, no existing system can perform this task without being explicitly trained which alters the quality of the answers. Common methods such as few-shot prompting and fine-tuning tend to modify the answer. Thus, we resort to a modified version of the SEMQA prompt {\cite{schuster2023semqa}, whereby instead of prompting the system to generate an answer via few-shot prompting, we instruct it to identify spans within a provided answer (prompt included in Appendix \ref{GPTprompt}, Figure \ref{fig:prompt_1}). We prompt GPT 3.5\footnote{\url{https://openai.com/blog/chatgpt}} and GPT-4\footnote{\url{https://openai.com/research/gpt-4}} using this modified prompt and they serve as our baselines for the first sub-task.

\subsubsection{Baselines for Sub-task 2}

GPT 3.5 and GPT-4 prompted with the modified SEMQA prompt (prompt included in Appendix \ref{GPTprompt}, Figure \ref{fig:prompt_2}) are used as baseline for the second sub-task as well, i.e., attributing the identified spans to their original source paragraph. In addition, we also employ retrieval methods such as BM25 (sparse) \cite{robertson2009probabilistic}, GTR (dense) \cite{ni2021large}, and MonoT5 \cite{nogueira2020document} as baselines for this task. 

By comparing the performance of our system against these baselines, we aim to evaluate its effectiveness  in tracing the origin of the spans.

\section{Results and Analysis}
We perform experiments for our approach (\S\ \ref{method}) with the following LLMs: Llama-7b \cite{touvron2023llama}, Llama-70b \cite{touvron2023llama}, Mistral-7b \cite{jiang2023mistral}, Yi-6b \cite{young2024yi} and OPT-350m \cite{zhang2022opt}. Note that OPT-350m has been excluded from comparision on \textsc{Verifiability-granular} dataset due to its limitation of 2048 token context length and the dataset contains samples upto 4096 tokens. All of our experiments were run on a A100 machine with 4 80GB GPUs. 
\subsection{Results for Sub-task 1}
The performance of our method and the baselines on sub-task 1, i.e., identifying the copied spans, is summarized in Table \ref{tab:sub-task1-modified} and Figure \ref{fig:sub-task1_veri} for QuoteSum and \textsc{Verifiability-granular} datasets respectively. We observe that our proposed method performs better than GPT models on the QuoteSum dataset and performs on par with GPT-4 on \textsc{Verifiability-granular}.

As seen in Table \ref{tab:sub-task1-modified}, while trying to optimize for F1 on the \textsc{Verifiability-Granular} dataset the scores we obtain indicate that the optimal solution is to mark all tokens as copied. Therefore, for a better analysis of model performance we plot the PR curves as shown in Figure \ref{fig:sub-task1_veri}.

\begin{table*}[h]
\centering
{\small 
\begin{tabular*}{\textwidth}{@{\extracolsep{\fill}}lcccccc@{}}
\toprule
\multirow{2}{*}{Model} & \multicolumn{3}{c}{QuoteSum} & \multicolumn{3}{c}{Verifiability-Granular} \\
\cmidrule(lr){2-4} \cmidrule(lr){5-7}
& P & R & F1 & P & R & F1 \\
\midrule
GPT-3.5 & 0.92 & 0.46 & 0.56 & 0.46 & 0.29 & 0.36 \\
GPT-4 & \textbf{0.96} & 0.87 & 0.90 & \textbf{0.76} & 0.83 & 0.79 \\
Llama-7b (Ours) & 0.96 & 0.97 & \textbf{0.96} & 0.73 & \textbf{0.99} & \textbf{0.84} \\
Mistral-7b (Ours) & 0.94 & 0.98 & \textbf{0.96} & 0.73 & 0.99 & 0.84 \\
Yi-6b (Ours) & 0.94 & \textbf{0.99} & 0.96 & 0.73 & 0.99 & 0.84 \\
OPT-350m (Ours) & 0.94 & 0.99 & 0.96 & - & - & - \\
\bottomrule
\end{tabular*}
}

\caption{Token level P, R \& F1 scores for identifying output tokens extracted from the document on QuoteSum and Verifiability test sets. For PR plot on \textsc{Verifiability-granular} dataset, please refer to Figure \ref{fig:sub-task1_veri}}
\label{tab:sub-task1-modified}
\end{table*}

The consistency in performance metrics across all models for this sub-task suggests that the capability to identify extracted tokens via hidden representations is not confined to any particular model family or size. This observation underscores the broader applicability and versatility of our proposed methodology, affirming its effectiveness regardless of the underlying architecture or capacity of the language model being used.

\begin{figure}[h]
    \centering
    \includegraphics[width=\linewidth]{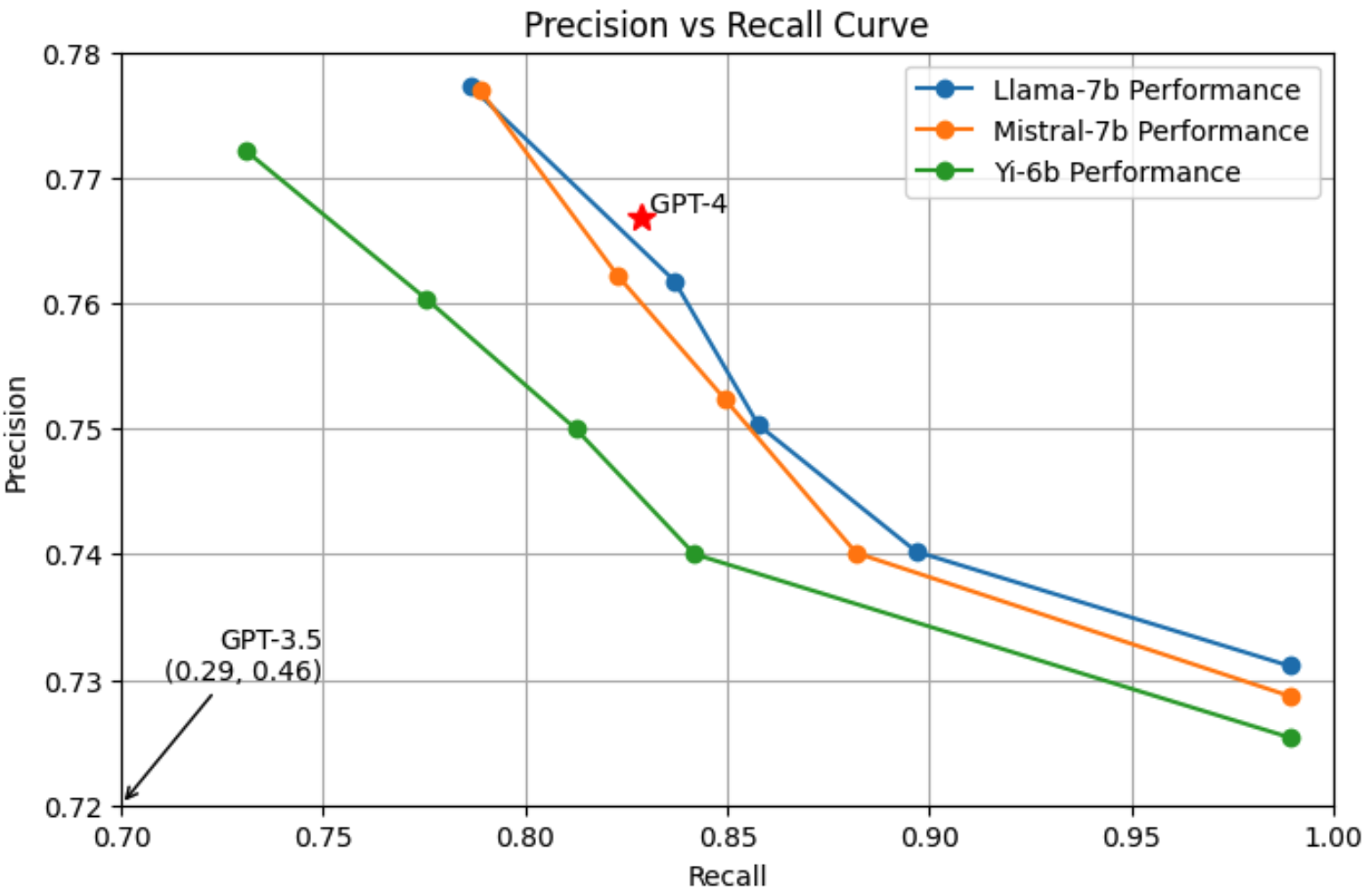}
    \caption{Performance of our method and baselines on the \textsc{Veri-gran} test set illustrated  using the Precision-Recall curve.}
    \label{fig:sub-task1_veri}
\end{figure}

The selection of the best hyperparameters (layer, threshold) for each of our models was based on the F1 performance metric. We share more details in the Appendix (Section \ref{appendix:hyperparameters}). The results presented in Table \ref{tab:sub-task1-modified} and Figure \ref{fig:sub-task1_veri} are derived from the test set.

The results indicate a considerable shortfall in the performance of GPT-3.5, particularly in terms of recall. This suggests that the model is unable to accurately identify all tokens considered by the annotators as being extracted from the document. A representative example of this failure case is illustrated in the Appendix (Figure \ref{fig:task1_failure}, Section \ref{appendix:limitations}).

As can be seen from the example, GPT-3.5 tends to identify entities rather than the specific spans that have been directly copied from the document. Even GPT-4, which demonstrates a better understanding of the task, occasionally overlooks certain components like "The" that are parts of the directly copied spans. 

An additional challenge with the GPT models is their tendency to hallucinate, leading to the introduction or elimination of content that doesn't exist in the original answer. We try to recover from the failure by taking spans marked on the modified response and superimposing it on the original response.

To elaborate our observations, we delve into the performance variation across different model families and layers in this sub-task, as illustrated in Figure \ref{fig:sub-task1}. For more precise comparisons among closely located numbers, the later layers from the Yi-6b model have been excluded due to their substantially lower performance compared to others.

The smoothest graph among all the model families belongs to OPT-350m which shows a gradual increase in performance until the middle layer, followed by a similarly gradual decrease. This performance peak at the middle layers aligns with the findings of \cite{zou2023representation}, suggesting that earlier layers are often dedicated to low-level tasks, while later layers tend to be excessively focused on next token prediction.

Interestingly, the larger models, despite their varied architectures, exhibit a shared performance trend, with a peak at the earlier layers followed by a subsequent decrease. This pattern suggests that for these larger models, the sub-task of identifying extracted tokens tends to be a relatively low-level or straightforward task. Consequently, the best performance for these models is typically observed in the earlier layers.

\begin{figure}
    \centering
    \includegraphics[width=\linewidth]{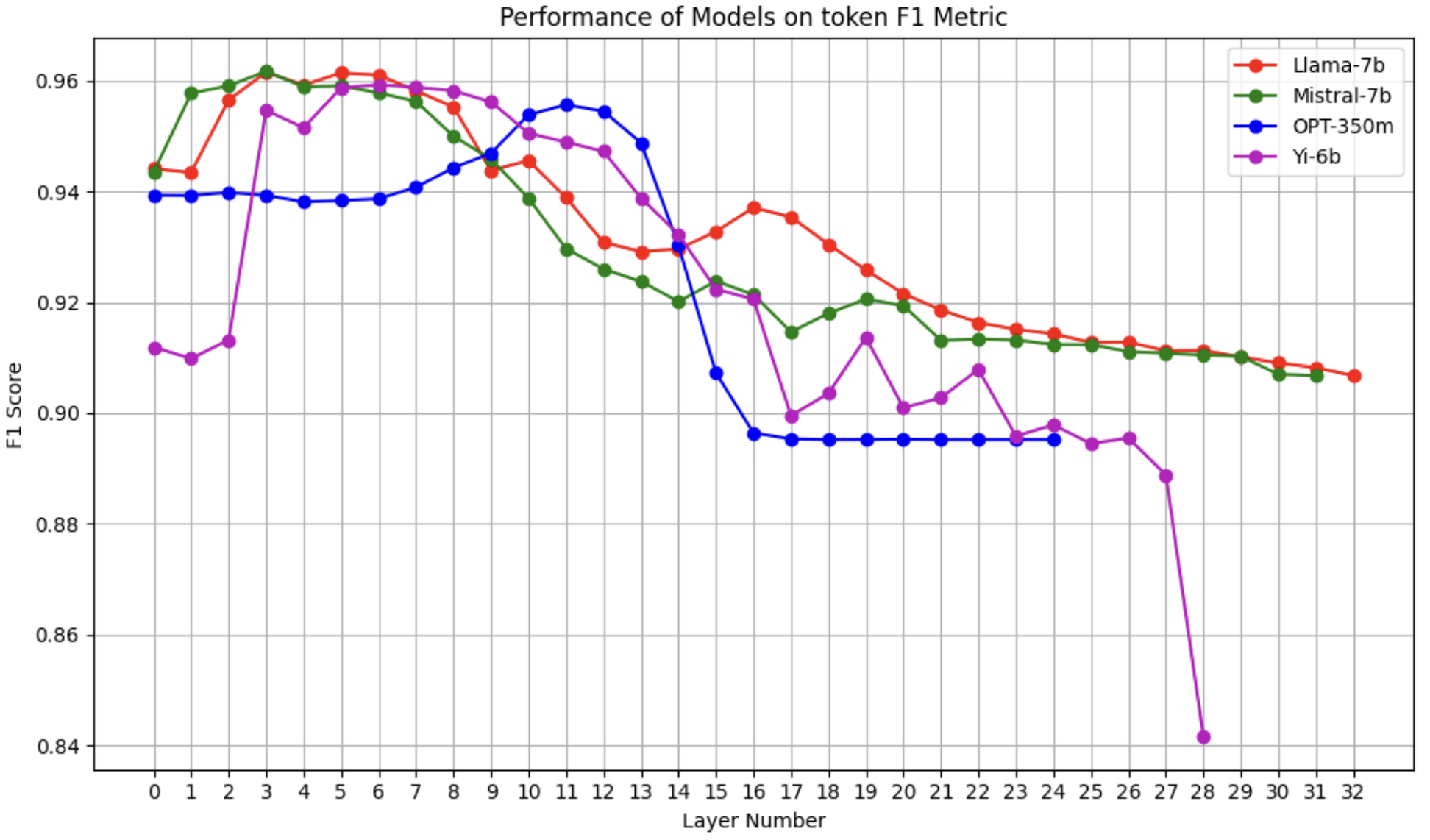}
    \caption{Comparison of model token F1 performance across layers of different models, for identifying output tokens extracted from the document on QuoteSum train set.}
    \label{fig:sub-task1}
\end{figure}

\subsection{Results for Sub-task 2}

Table \ref{tab:sub-task2} presents the performance results for the second sub-task. For each model utilizing our methodology in this task, the sole hyperparameter is the layer. The results provided in the table are derived from evaluations conducted on the test set of QuoteSum and \textsc{Verifiability-granular} datasets. 

\begin{table}[h]
    \centering{
    \small
    \begin{tabular}{@{}lcccc@{}}
        \toprule
        Model & \multicolumn{2}{c}{Accuracy (\%)} \\
        & \textsc{QuoteSum} & \textsc{veri-gran} \\
        \midrule
        GPT-3.5 & 90.18 & 26.40 \\
        GPT-4 & \textbf{90.59} & 62.11 \\
        \midrule
        BM25 & 75.72 & 68.20 \\
        GTR & 72.57 & 53.15 \\
        MT5 & 89.24 & 67.43 \\
        \midrule
        Llama-7b (Ours) & 87.51 & 77.33 \\
        Mistral-7b (Ours) & 89.95 & \textbf{77.71} \\
        Yi-6b (Ours) & 89.24 & 77.61 \\
        OPT-350m (Ours) & 75.29 & -- \\
        \bottomrule
    \end{tabular}
    }
    \caption{Paragraph-level accuracy for attributing extractive spans on QuoteSum and \textsc{Veri-gran} test sets.}
    \label{tab:sub-task2}
\end{table}

The results indicate that our methods perform on par with GPT-4 ($\sim 90\%$) on the QuoteSum dataset, while they outperform GPT-4 by $\sim 15\%$ on \textsc{Verifiability-granular}. Note that the human performance on \textsc{Veri-gran} is 92.04\%, calculated as the mean performance of 4 NLP practitioners on the dataset.

The reason for this is the much harder nature of the later dataset due to the larger number of passages (3.38 vs 69.1 passages on average). Prompting GPT to attribute already generated answers is essentially a retrieval task. It has been observed by \cite{liu2023lost} that LLM retrieval performance degrades with increasing context length. Our method benefits from the contextual representations of LLM embeddings without relying on the LLM for retrieval and matching representations directly.

Two noteworthy observations from the dataset and results are:

\begin{enumerate}
    \item The specific nature of the datasets necessitate a delicate balance between two competing tasks: directly matching the substring in the span to be attributed, and disambiguating among multiple instances of the substring. The former is a low-level task, while the latter requires a more nuanced, high-level understanding of context-based matching.
    \item The successful performance of models across different families and capacities suggests that approach of matching based on hidden representations is broadly applicable.
\end{enumerate}

Similar to sub-task 1, we carry out a detailed analysis of performance on sub-task 2 across various model configurations and layers, as depicted in Figure \ref{fig:sub-task2}.

A fundamental difference between the OPT-350m and larger models lies in the influence of token positions on the layer 0 embedding. For OPT-350m, the token's position directly impacts its layer 0 representation. This differs significantly from larger models, where the layer 0 representation depends solely on the token ID like Word2Vec \cite{mikolov2013efficient}, with the token's position not having a direct influence. Instead, in larger models, the token position is a part of the transformer's learning process \cite{su2023roformer}.

In light of this distinction, our methodology, observed in Table \ref{tab:sub-task2}, excludes layer 0 for the larger models while choosing the best layer. Because its representation, which is solely dependent on the token ID, tends to prioritize exact substring matches. 

Therefore, OPT-350m begins with lower performance that gradually at the middle and the final layers. On the other hand, all larger models across different families exhibit a similar trend where they start with the highest performance at layer 0. This high initial performance is largely attributable to the dataset's characteristics, which contain numerous examples that can be unambiguously matched using exact substring match.

However, the dataset also comprises examples with multiple similar substrings, necessitating disambiguation. This requirement for contextual understanding is well catered to by the middle and later layers, as our ablation study in the Appendix (Section \ref{appendix:disambiguation}) reveals.
Thus, the slight upturns in performance experienced by the larger models in the middle and later layers indicates the trade-off between exact and contextual matching.

\begin{figure}
    \centering
    \includegraphics[width=1\linewidth]{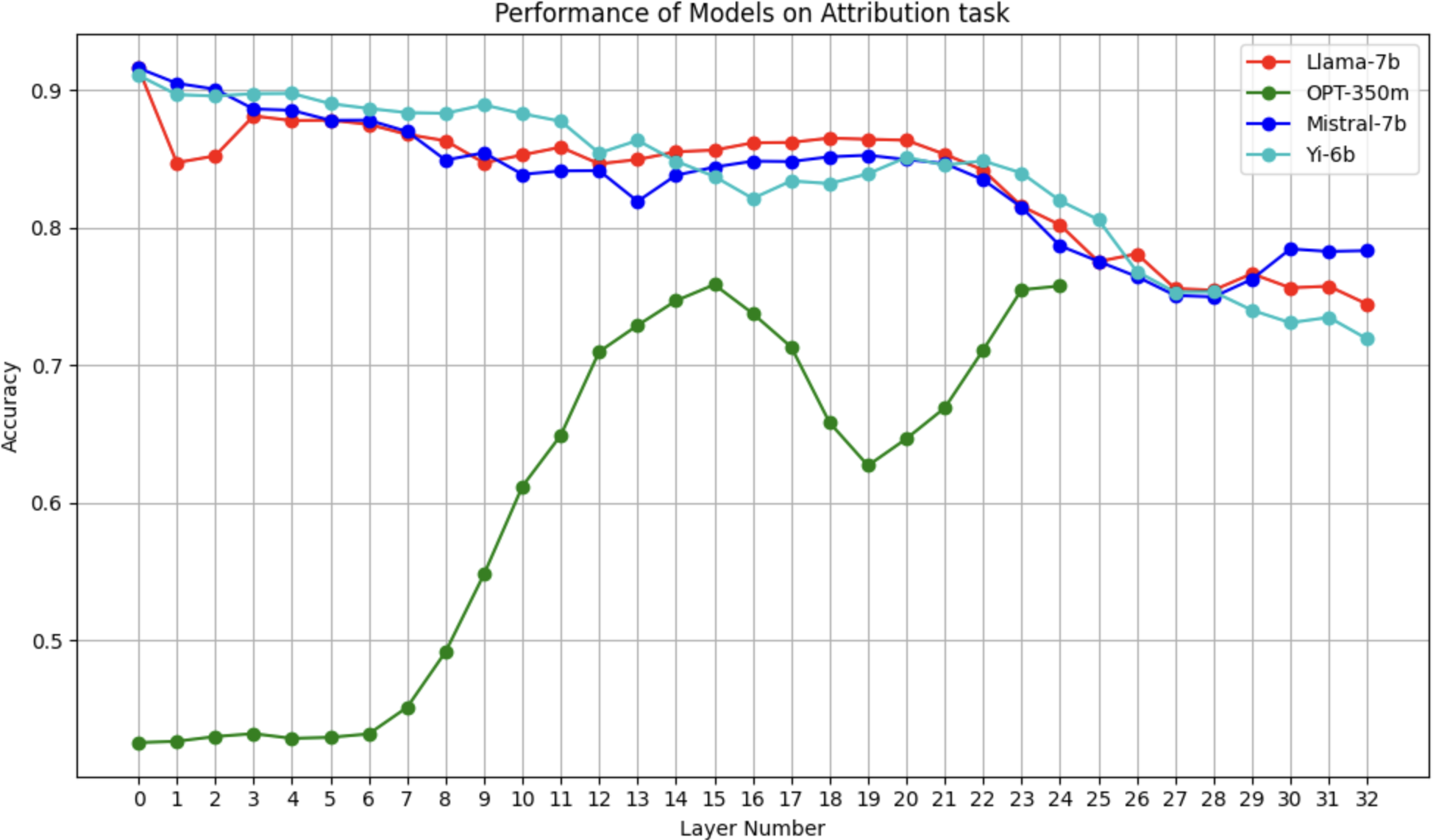}
    \caption{Comparison of model accuracy across layers for attributing extractive spans on QuoteSum train set.}
    \label{fig:sub-task2}
\end{figure}

\section{Discussion \& Conclusion}

Through this study, we introduce a novel, efficient solution to the challenge of attribution in contextual question answering. Our method leverages the hidden state representations of LLMs, providing detailed, granular attributions without requiring extensive model retraining.
Our approach's ability to perform token-level attribution easily lends itself to end-user applications, and we have included an actual system input-output in the Appendix (Section \ref{appendix:qualitative_examples}) depicting the same.

Our experimental results, across various model configurations and layers, demonstrate the efficacy of our approach. Notably, our method performs on par or better than training-free baselines in both identifying tokens in an answer that are directly copied from the context and attributing these tokens to their original positions in the document. Additionally we make available \textsc{Verifiability-granular} dataset, which contains token-level attributions for LLM generations in the contextual question answering setup.

Perhaps one of the most interesting findings from our study is that the ability to identify extracted tokens and attribute them back to the source is not exclusive to a specific model family. This suggests that our approach can be applied broadly across different LLM architectures, underscoring its versatility.

\section{Limitations \& Future Work} 

One limitation of our work stems from the specific nature of the QuoteSum and \textsc{Verifiability-granular} datasets we utilized for our experiments. The datasets are constructed such that only verbatim spans from the document are annotated and attributed in the answers. Our method, thus, is primarily evaluated on verbatim spans. However, our method is not specifically designed for verbatim spans only; it could potentially work with paraphrased information as indicated in our experiments with a synthetically paraphrased version of Quotesum presented in the Appendix (Section \ref{appendix:paraphrase_exp}).

In the future, we intend to employ datasets that include paraphrased spans. Testing on such data will allow us to assess the method's effectiveness in attributing paraphrased information, identify potential challenges, and adapt our methodology accordingly.

 We also see potential for our method to be useful beyond the realm of contextual question answering. We tried mapping an LLM generated text span in Spanish to its source document in English (Figure \ref{fig:spanish_ex}). 

 \appendix
 
 Interestingly, our method shows potential for this use case too. We envision applying our approach to other tasks that involve information extraction and attribution to a given context. This will not only enhance our understanding of the generation process of LLMs but also extend the applicability and value of our method. We are currently limited  by the availability of datasets to perform detailed analyses of other applications.

\bibliography{anthology,custom}
\bibliographystyle{acl_natbib}

\appendix
\begin{figure*}
    \centering
    \includegraphics[width=1\textwidth]{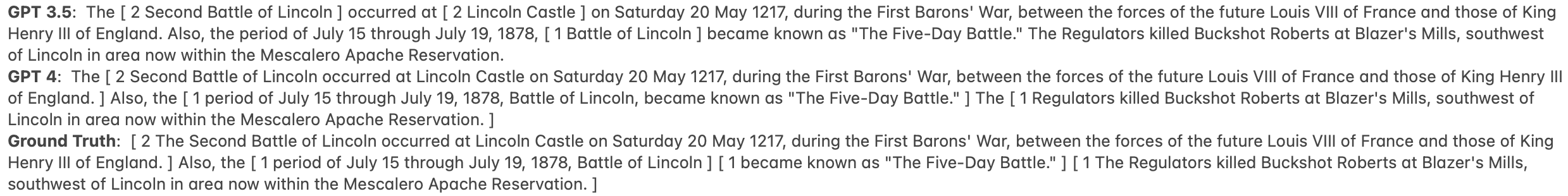}
    \caption{Limitations of GPT models in identifying directly copied spans from a source document}
    \label{fig:task1_failure}
\end{figure*}

\begin{figure*}[h]

\centering
\includegraphics[width=.32\textwidth]{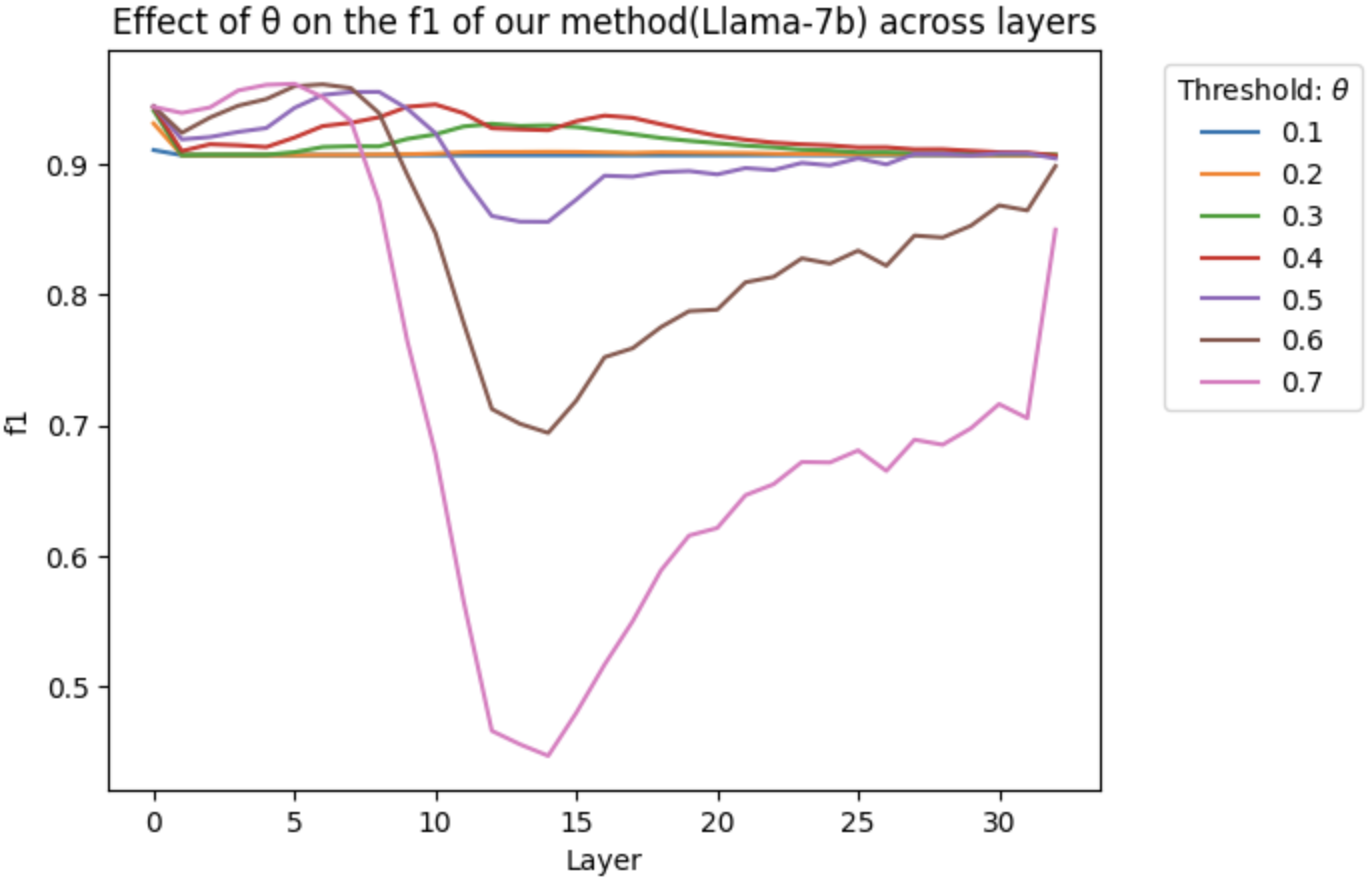}\hfill
\includegraphics[width=.32\textwidth]{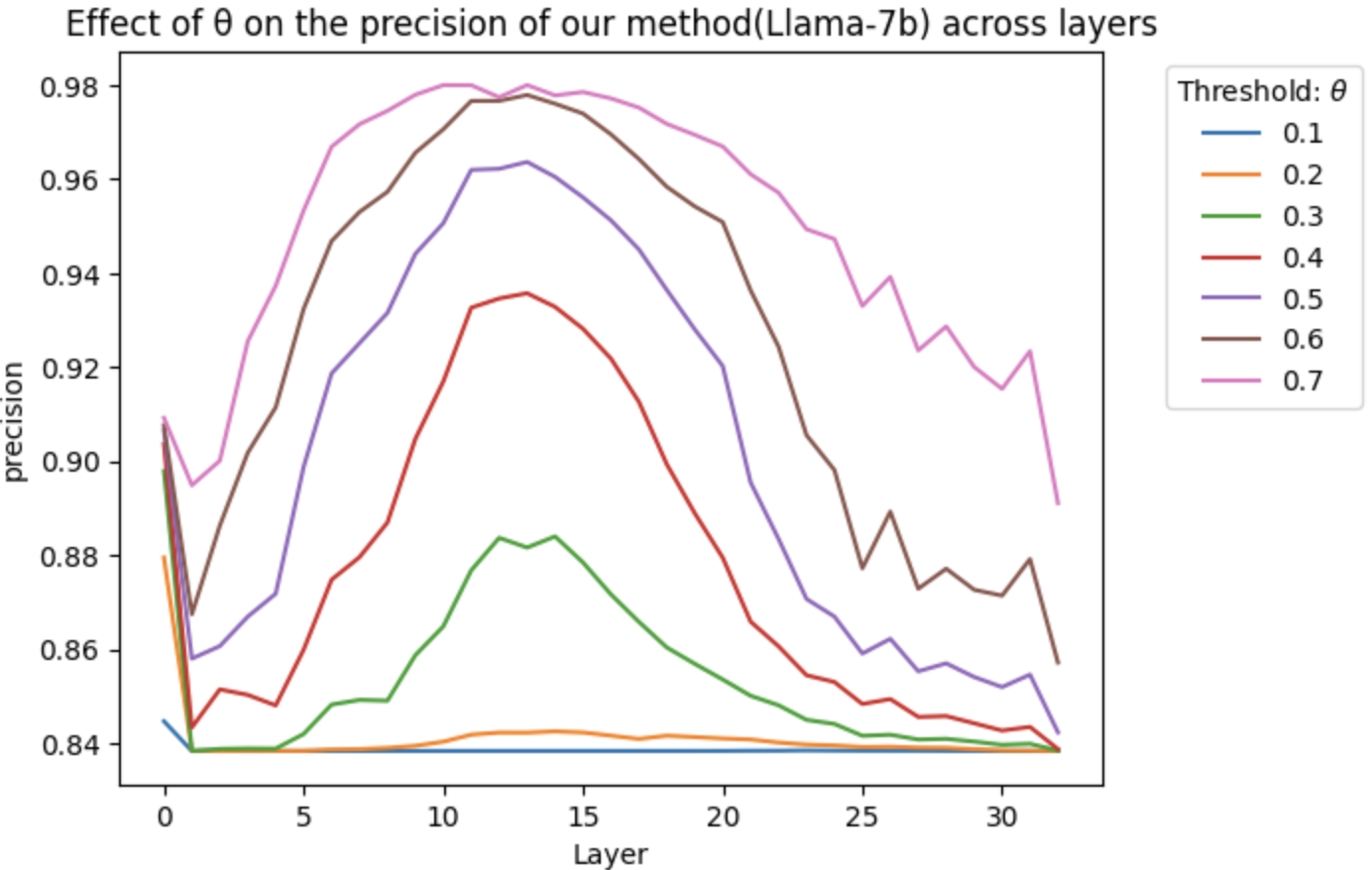}\hfill
\includegraphics[width=.32\textwidth]{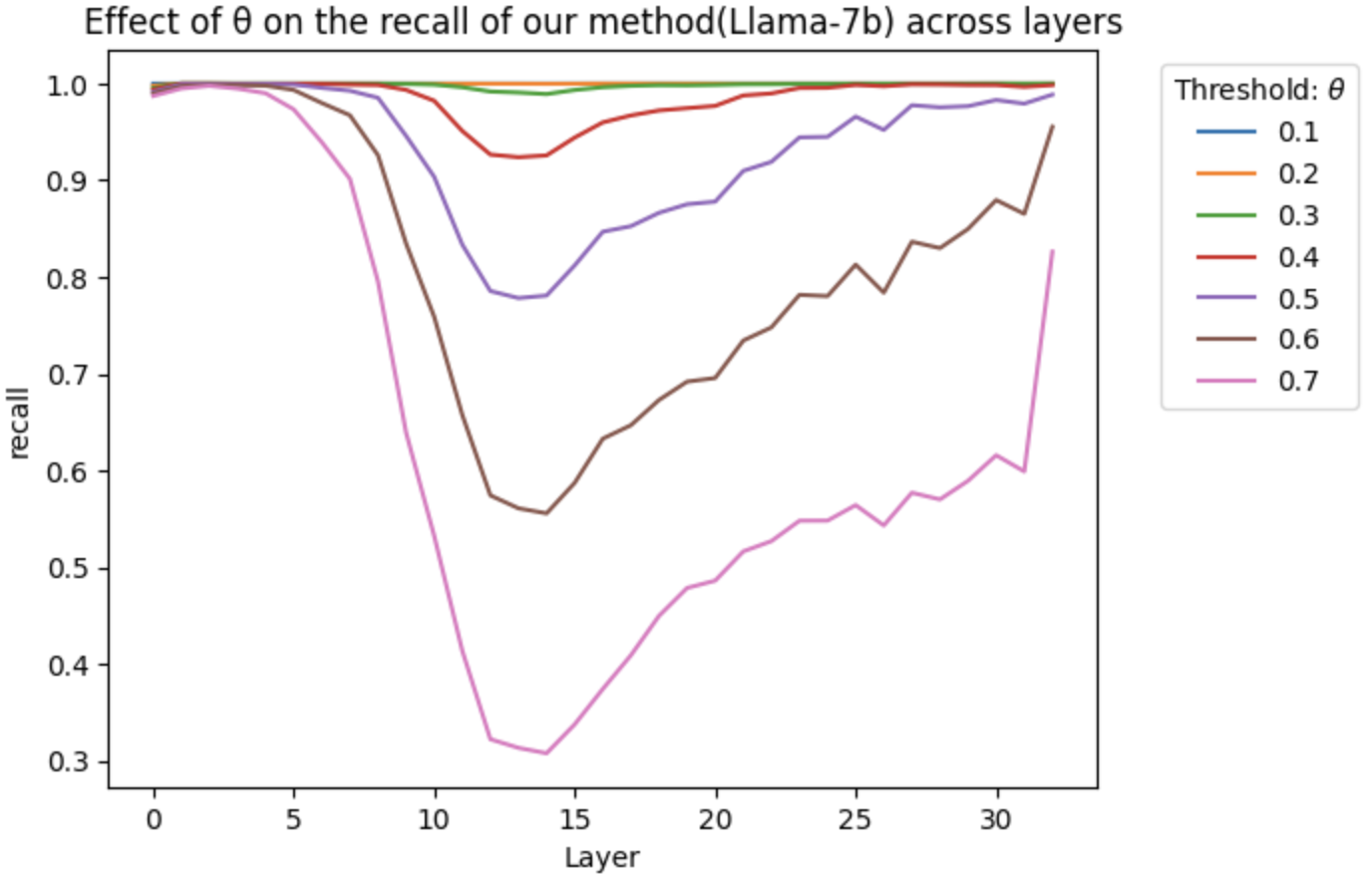}

\caption{Effect of the threshold \textsc{$\theta$} on the performance of our method across different layers of Llama-7b on QuoteSum train set. }
\label{fig:hyp}

\end{figure*}

\section{Appendix}
\subsection{Performance over target span position}
In this section, we analyze if the position of the target span (the span that needs to be attributed) within the answer influences the efficacy of our methodology. The start character of each span is normalized with respect to the length of the answer and the accuracy is measured for all the models on Sub-task2. As shown in Figure \ref{fig:span_position}, the accuracy improves with increasing span position for all models. 
In decoder-only models each step can access information from previously generated or processed tokens only. As the model progresses through the text, it accumulates more information and context about the text it is analyzing. We believe that this behaviour is the reason why accuracy increases when we attribute later text spans.
\label{appendix:target_span_pos}

\begin{figure}[h]
    \centering
    \includegraphics[width=1\linewidth]{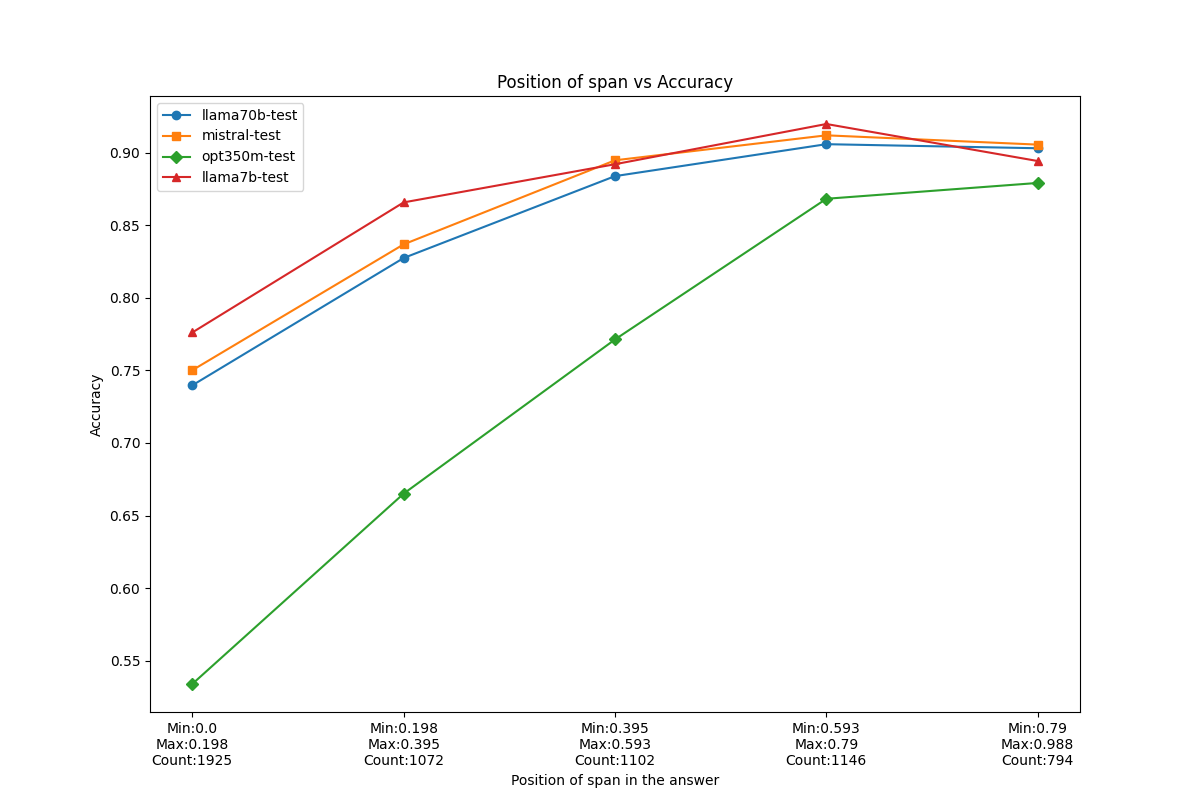}
    \caption{Performance of different models over span position on Sub-task 2.}
    \label{fig:span_position}
\end{figure}

\subsection{Performance of layers in the disambiguation task setting}

Given a span that needs to be attributed, there are instances in the dataset where ground truth attribution sub-string is present multiple times over different sources. During inference, our approach has to both identify the correct attribution sub-string and also disambiguate between the multiple occurances of the sub-string. Disambiguating over these multiple occurences, requires contextual understanding of the span to be attributed in order to choose between the occurences. Data points that require disambiguation are collected from the test set of the QuoteSum dataset. 

We measure performance across all layers for different models on this subset (Figure \ref{fig:diasmb_task}). We compute the performance when one of the multiple occurrences of the sub-string is chosen randomly. This is highlighted as the random baseline in the figure.

Figure \ref{fig:diasmb_task} shows an interesting trend, where early layers perform worse than the random baseline on the disambiguation task. Early layers capture low level, basic features of the input text, making them less suitable for a complex task such as disambiguation. Performance improves as we use later layers, with the middle layers yielding the best performance for Llama-70b and Mistral-7b, and later layers yielding the best performance for Llama-7b and OPT-350m.
\begin{figure}
    \centering
    \includegraphics[width=1\linewidth]{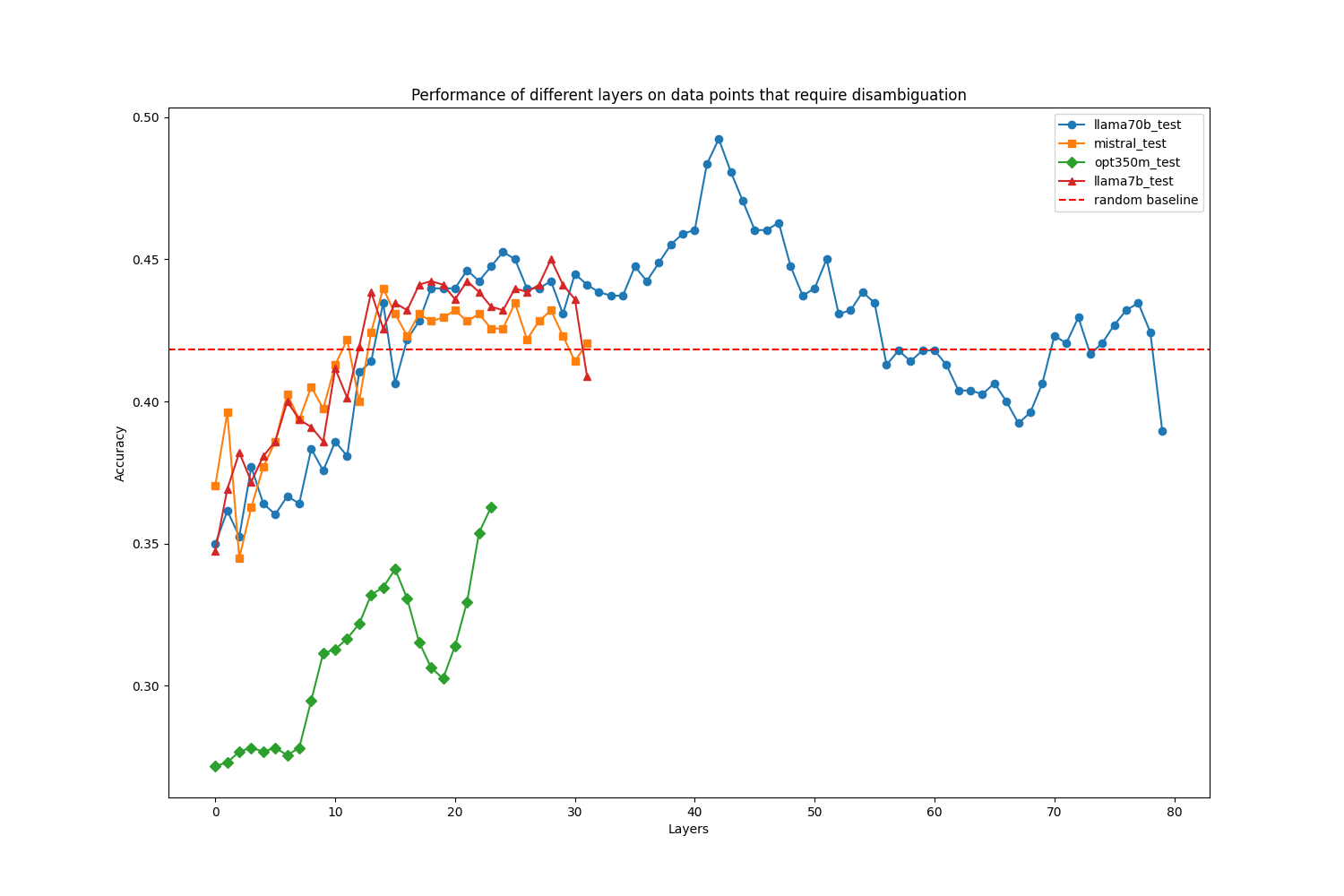}
    \caption{Performance of different models over the disambiguation task.}
    \label{fig:diasmb_task}
\end{figure}
\label{appendix:disambiguation}

\subsection{Effect of hyper-parameters on Sub-task 1 performance}
We present graphs for precision, recall and F1 in Figure \ref{fig:hyp} generated during our experiments on the QuoteSum train set, which show the impact of the threshold $\theta$, and the choice of layer on the performance of our method (Llama-7b) for Sub-task 1. $\theta$ was chosen for each layer-model setup to achieve the best f1 score. Interestingly we note that in the earlier layers, higher thresholds are preferred. We hypothesize that there's a higher chance of low-level information overlapping, necessitating more stringent filtering at the earlier layers.
\label{appendix:hyperparameters}

\subsection{Limitations of GPT models for Sub-task 1}
Figure \ref{fig:task1_failure} depicts a qualitative example where GPT models fail in identifying directly copied spans from a source document on the QuoteSum dataset. In this example GPT-3.5 is only attributing entities like Lincoln Castle, Battle of Lincoln and the Second Battle of Lincoln. GPT-4 has a better understanding of the task but leaves out tokens like 'The' which were copied verbatim from the source. 
\label{appendix:limitations}

\subsection{Synthetic Paraphrasing of QuoteSum}
\label{appendix:paraphrase_exp}
We paraphrase the QuoteSum dataset by prompting GPT-4 as shown below. A sample from the dataset is shown in Figure \ref{fig:paraphrased_sample}.

\begin{figure*}
    \centering
    \includegraphics[width=1\textwidth]{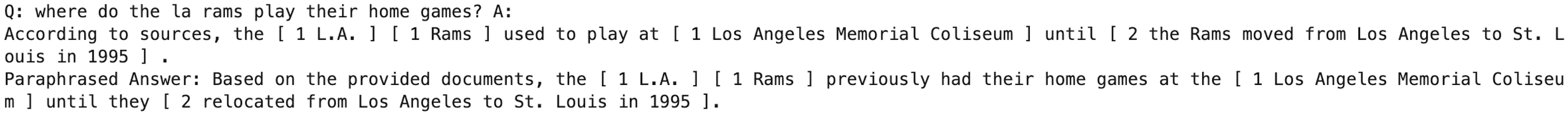}
    \caption{Sample datapoint from the paraphrased QuoteSum dataset}
    \label{fig:paraphrased_sample}
\end{figure*}

\begin{lstlisting}[basicstyle=\small\ttfamily, breaklines=true, breakatwhitespace=true]
Document:
{document}
Based on the information contained in the document, answer the question with details to the best of your abilities. Think step by step and explain your answer if that will help better understand the answer. 
Q: {question} A:
{answer}
Given the above source passages, a question and an answer. The answer summarizes the given sources while explicitly copying spans from the sources. Paraphrase the non-entity parts of the answer within [] while keeping entities intact and rewrite in the same format as original answer.
Paraphrased Answer: 
\end{lstlisting}
We computed the sub-task 2 performance of the various methods on the paraphrased test set and summarize the results in Table \ref{tab:paraphrased_quotesum}.

\begin{table}[h]
    \centering
    \begin{tabular}{@{}lccc@{}}
        \toprule
        Model & Accuracy \\
        \midrule
        BM25 & 73.99 \\
        GTR &  71.79 \\
        MT5 &  \textbf{88.46} \\
        \midrule
        Llama-7b (Ours) & 85.74 \\
        Mistral-7b (Ours) & 87.06 \\
        Yi-6b (Ours) & 86.99  \\
        OPT-350m (Ours) & 71.18 \\
        \bottomrule
    \end{tabular}
    \caption{Paragraph-level accuracy for attributing extractive spans on paraphrased QuoteSum test set.}
    \label{tab:paraphrased_quotesum}
\end{table}

We observe that the performance of all the methods drop on the paraphrased version of QuoteSum by nearly the same amount(1-3\%). This probably implying that synthetic paraphrasing setup while harder than the original dataset is not difficult enough to offer additional insights.

\subsection{Qualitative Example for Sub-task 1 and Sub-task 2}

To illustrate the utility of our method, a qualitative example is presented in Figure \ref{fig:task1_openai}.
\label{appendix:qualitative_examples}
\begin{figure*}[h]
    \centering
    \includegraphics[width=1\textwidth]{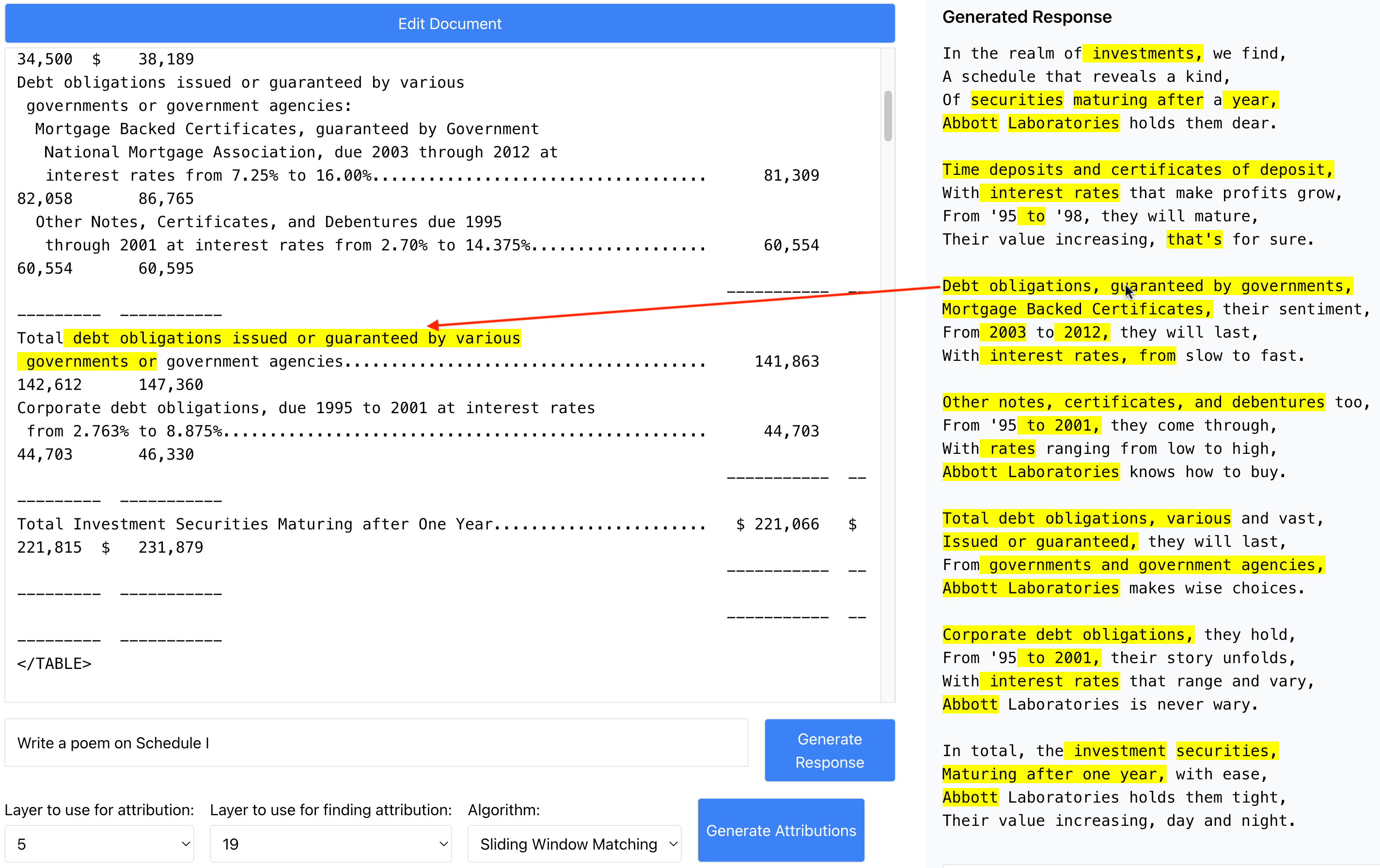}
    \caption{Given a document and a question, the user selects generate response to get the answer. Once the answer is generated, to identify extractive answer tokens, the user selects the Generate attributions button. Answer tokens copied from the document are highlighted and are clickable. When the user selects 'Debt Obligations..' in the generated response, the corresponding attribution is shown on the left panel.}
    \label{fig:task1_openai}
\end{figure*}

\begin{figure*}[h]
    \centering
    \includegraphics[width=1\textwidth]{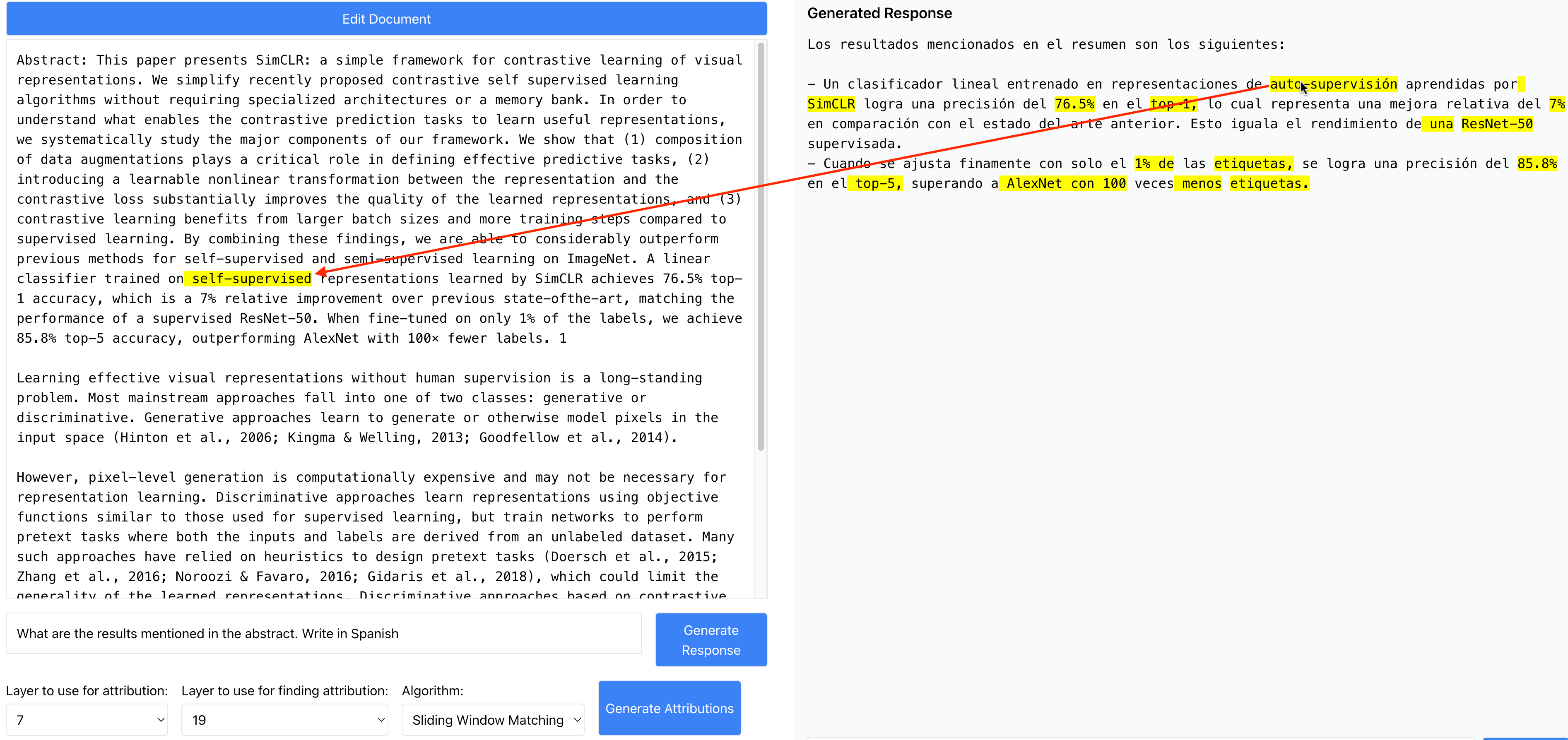}
    \caption{Given a document and a question, the user selects "Generate response" to get the answer. Once the answer
is generated, to identify extractive answer tokens, the user selects the "Generate Attributions" button. Answer tokens
copied from the document are highlighted and are clickable. When the user selects "auto super.." in the
generated response, the corresponding attribution is shown on the left panel.
}
    \label{fig:spanish_ex}
\end{figure*}

\subsection{Prompts for GPT baselines}
\label{GPTprompt}
We present the prompts used for our GPT-based baselines for Sub-task 1 and Sub-task 2 in Figure \ref{fig:prompt_1} and Figure \ref{fig:prompt_2} respectively.
For Sub-task 1, the LLM has to respond with copied spans along with the source paragraph number. For Sub-task 2, we mark the verbatim copied spans in "[]" and prompt the LLM to identify the source paragraph only. 
\begin{figure*}[h]
    \centering
    \includegraphics[width=1\textwidth]{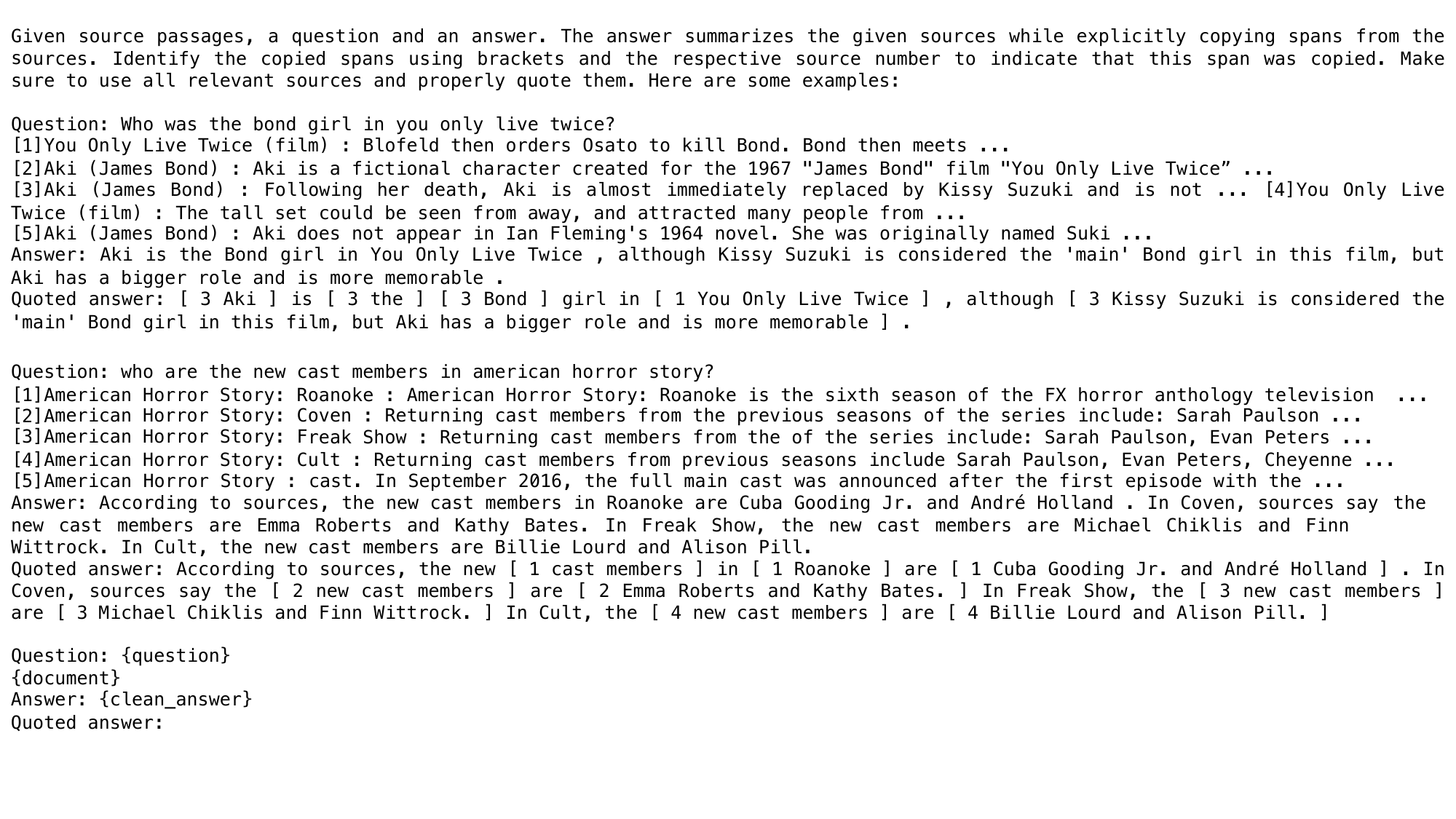}
    \caption{Prompt for GPT-based baselines for Task-1.}
    \label{fig:prompt_1}
\end{figure*}
\begin{figure*}[h]
    \centering
    \includegraphics[width=1\textwidth]{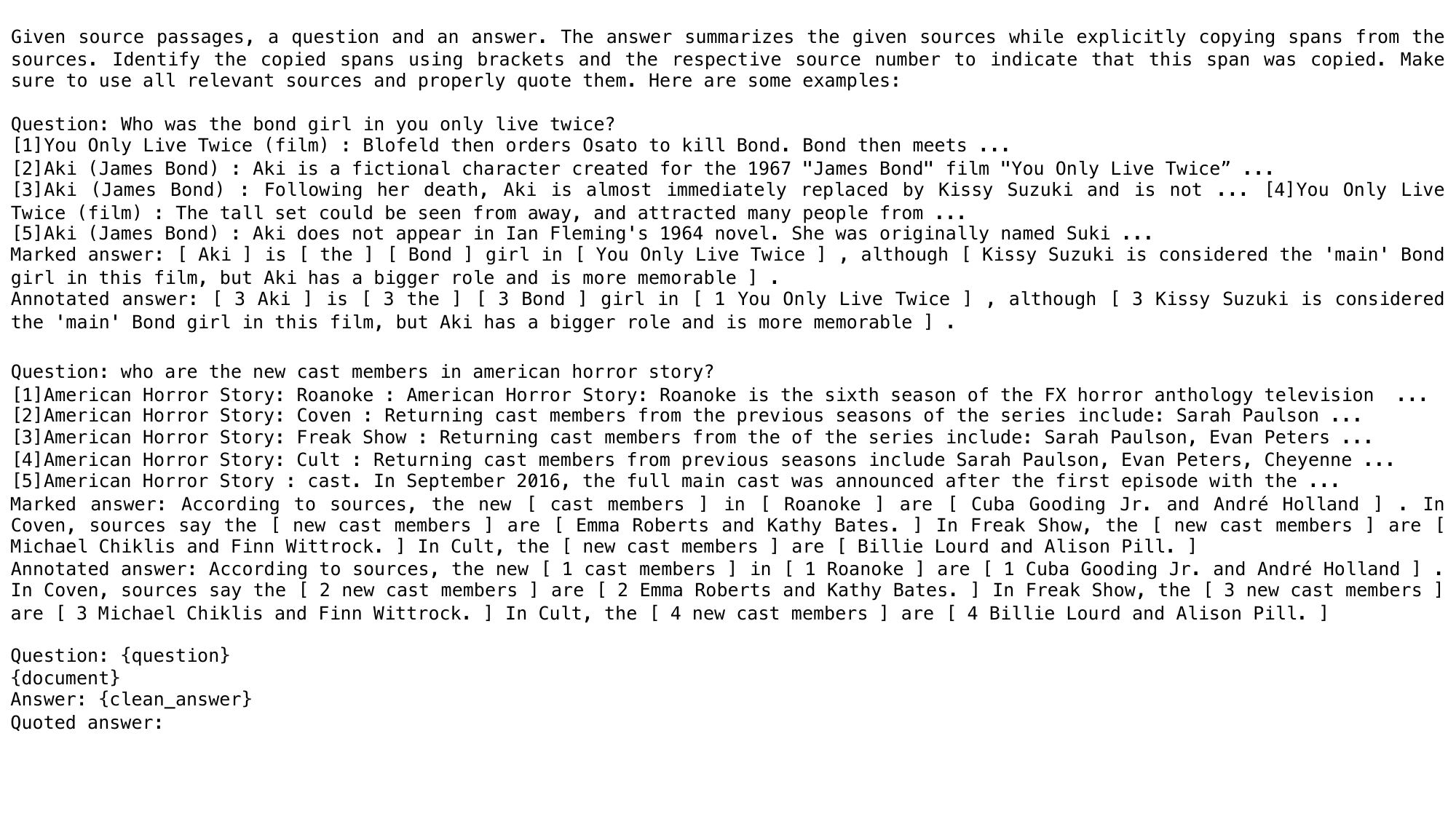}
    \caption{Prompt for GPT-based baselines for Task-2.}
    \label{fig:prompt_2}
\end{figure*}

\end{document}